%% file: darwin-thesis.tex
\documentclass[11pt]{thesis}
\usepackage{epsfig}
\begin{document}

%% Switch page numbering to Roman numerals and
%% suppress the numbering of chapters
\frontmatter
\thispagestyle{empty}
\setlength{\parindent}{0.0in}
\begin{center}
{\large
Unsupervised Thresholding for Automatic Extraction of Dolphin Dorsal Fin Outlines from Digital Photographs in DARWIN

(Digital Analysis and Recognition of Whale Images on a Network)

\vspace{0.75in}

                                    by

                                    Scott A. Hale

\vspace{1.125in}

                   A Senior Thesis in Computer Science\\
           submitted in partial fulfillment of the requirements\\
                            for the degree of

\vspace{0.5in}

                           Bachelor of Science

\vspace{0.5in}

                                    at

\vspace{0.75in}

                              Eckerd College
                              
                         St. Petersburg, Florida

\vspace{1.0in}

                               May 18, 2008
}
\end{center}

\newpage
\thispagestyle{empty}%supress page number
\mbox{}%something just so latex will actually add another page
\newpage
\thispagestyle{empty}

\begin{center}
{\large
                             COMMITTEE REPORT
}
\end{center}

\vspace{0.8in}

A Thesis in Computer Science.

\renewcommand{\baselinestretch}{0.5}
This Thesis was prepared under the direction of the Research Advisor and the candidate's
Thesis Committee.  It was submitted to the Faculty of Eckerd College and was approved as
partial fulfillment of the requirements for the degree of Bachelor of Science.
\renewcommand{\baselinestretch}{2.0}

\vspace{0.875in}

May 18, 2008

\vspace{0.8in}

\hspace{2.32in}
                              THESIS COMMITTEE:

\renewcommand{\arraystretch}{0.25}
\begin{tabular}[h]{lll}
\vspace{0.8in} \\ \cline{3-3}\\
& \hspace{2.0in} & Kelly R. Debure, Research Advisor\\
\vspace{0.8in} \\ \cline{3-3}\\
& \hspace{2.0in} & Edmund Gallizzi\\
\vspace{0.8in} \\ \cline{3-3}\\
& \hspace{2.0in} & Shannon Gowans\\
\end{tabular}

\setlength{\parindent}{0.4in}
\begin{abstract}
\hspace{0.4in}
At least two software packages---DARWIN, Eckerd College \cite{darwin}, and FinScan, Texas A\&M \cite{finscan}---exist to facilitate the identification of cetaceans---whales, dolphins, porpoises---based upon the naturally occurring features along the edges of their dorsal fins. Such identification is useful for biological studies of population, social interaction, migration, etc. The process whereby fin outlines are extracted in current fin-recognition software packages is manually intensive and represents a major user input bottleneck: it is both time consuming and visually fatiguing. This research aims to develop automated methods (employing unsupervised thresholding and morphological processing techniques) to extract cetacean dorsal fin outlines from digital photographs thereby reducing manual user input. Ideally, automatic outline generation will improve the overall user experience and improve the ability of the software to correctly identify cetaceans.
Various transformations from color to gray space were examined to determine which produced a grayscale image in which a suitable threshold could be easily identified. To assist with unsupervised thresholding, a new metric was developed to evaluate the jaggedness of figures (``pixelarity'') in an image after thresholding. The metric indicates how cleanly a threshold segments background and foreground elements and hence provides a good measure of the quality of a given threshold. This research results in successful extractions in roughly 93\% of images, and significantly reduces user-input time.

\end{abstract}

\pagebreak

\hspace{0.01in}
\vspace{0.01in}
\begin{center}
{\huge
{\bfseries
Acknowledgments
}
}
\end{center}
This research was supported by the National Science Foundation under
grant number DBI{}-0445126.

The author would like to thank his tireless thesis committee for their assistance.

\tableofcontents
\listoffigures%added by SAH, I just think it would be cool.

\renewcommand{\textfraction}{0.0}
\mainmatter
\include{chapter1}
\include{chapter2}
\include{chapter3}
\include{chapter4}

\include{chapter5}

\include{chapter6}

\include{bibliography}
\backmatter
\clearpage
\pagebreak

\end {document}

%% file: chapter1.tex
\chapter{Problem and Motivation}
\hspace{0.4in}

As cetaceans{---}whales, dolphins, porpoises{---}can be uniquely
identified by the features of their dorsal fins, researchers frequently
employ photo{}-identification techniques in studies of population,
migration, social interaction, etc. Researchers construct a catalog of
known individuals and classify fins based on the location of primary
identifying damage features. The subsequent comparison of unknown fins
to this catalog is time consuming and tedious. With the pervasive use
of digital cameras in such research, the quantity of photographic data
is increasing and the backlog of individual identification delays
meaningful data analysis. 

Automated fin comparison can potentially improve productivity for photo
identification tasks. Automated methods to compare dorsal fin
photographs exist{---}DARWIN \cite{darwin} and FinScan \cite{finscan}{---}and can
significantly reduce the number of photographs that must be examined
manually to correctly identify an individual. Using fin outlines for comparison,
these methods identify those fins from a database of known fins
which most closely match the new, unknown fin and display the results as
a ranked list. However, these methods
may require more work on the end user's part than
simply looking through a catalog of photos. Tracing a fin outline
accounts for a large portion of the time required for data input. Not
only is the process time consuming, but also the nature of the task
causes visual fatigue. Many institutions employ a digital catalog, but
do not employ an automated{}-recognition package. In fact, some
institutions still prefer the classic catalog approach to automated
methods.

In order to make automated fin comparison packages more usable and improve the end user
experience, a process to automatically extract a fin outline from a
photograph is desirable. Such a process would decrease both user input
time and the visual fatigue caused by manual outline extraction. As a
result, automated fin comparison packages would be more
user{}-friendly.

%% file: chapter2.tex
\chapter{Background and Related Work}

\begin{figure}[h]
\centerline
{\epsfxsize=4.5in\epsfbox{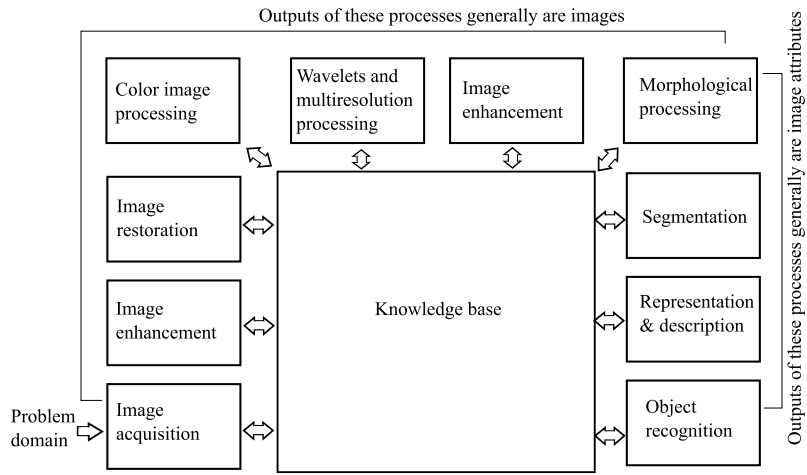}}
\begin{center}
\parbox{4.5in}
%% this goes into the figure list (which is just after the table of contents..)
{\caption[Fundamental Steps of Image Processing]
%% this goes under the picture
{\em Gonzalez \cite{gonzalez} describes several fundamental steps in digital image processing.}
\footnotesize

\label{fundamental_steps}}
\end{center}
\end{figure}

\hspace{0.4in}
Digital image processing is the application of computer algorithms to
digital images. There are two principal application areas of image processing: improving pictorial information for human interpretation
and processing scene data for autonomous machine perception \cite{gonzalez}. While the output of the first category is often images, the second category often aims to extract information from an image in a form suitable for computer processing \cite{gonzalez}. This research falls into the second category. The goal is to move from a pictorial description of a fin to a mathematical plot of points along the outline of this fin. Gonzalez \cite{gonzalez} identifies several fundamental steps in digital image processing (See Figure \ref{fundamental_steps}). The work described in this thesis focuses upon color image processing, segmentation, and morphological processing with the goal of representing and describing a dorsal fin by a plot of coordinate pairs marking its outline. Each of these steps draws upon a knowledge base of information about the problem domain. Ideally, this knowledge base will be as small as possible.

%The goal of image processing is to apply these algorithms in order to solve problems
%brief history of image processing

%classic techniques

 %histogram analysis{---}e.g. factory q.a., this differs b/c conditions:
%lighting, distance, etc. unknown/variable
Image processing applications vary greatly in success depending upon the conditions under which images are acquired. Images acquired under controllable conditions, for example, differ from images
acquired in the field. Images acquired in factories for quality assurance applications often have known lighting and distance parameters. This \emph{a priori} knowledge greatly simplifies the image
processing task and provides good consistency from image to image. For images acquired in the field,
such as those used in this research, lighting conditions and camera distance and angle are often not known and vary from one image to another. Ideally, the methods developed here will operate on as wide a domain of images and under as many different conditions as possible.

 %binary image operations

% color models

%new application

%pixelarity metric=new technique

% neighborhood approach (thin, erode, dilate) established by _________

\section{Histogram Analysis Techniques for Segmentation \\by Thresholding}
\hspace{0.4in}

Segmentation is the process by which regions in an image are classified into categories based upon similarities to and differences from surrounding regions. Commonly, segmentation is used to separate foreground objects from background regions in an image. This process results in a binary image in which the foreground is one value and the background another. Thresholding is a popular image segmentation technique \cite{sahoo} that involves analysis of an image's histogram. Thresholding techniques are those that determine the threshold value, $t$, based on certain criteria. All pixels with values less than or equal to $t$ are given one color while those greater than $t$ are given another.

Thresholding methods are either global or local and point{}- or region{}-dependent \cite{sahoo}. Global thresholding algorithms choose one threshold for the entire image while local thresholding algorithms partition the image into subimages and select a threshold for each subimage. Point{}-dependent thresholding algorithms only analyze the gray level distribution of the image while region{}-dependent algorithms also consider the location of the pixels.

The p{}-title method is one of the earliest thresholding methods \cite{sahoo}. The method requires an \emph{a priori} knowledge of the approximate percentage of pixels in the foreground object. A threshold $t$ is then selected so that at least this percentage of pixels is selected for the foreground. When the distance of the camera from the photographed object is not known, the resulting size of the desired object cannot easily be determined and the use of this method becomes problematic.

The mode method works well for images that are almost bimodal, having distinct foreground objects and background regions. The threshold chosen corresponds to the valley of the histogram \cite{sahoo}.

Where images are not bimodal and percentages are not known \emph{a priori}, it is difficult to find a good valley in the histogram upon which to threshold. In this case, it is often possible to define a threshold at the \emph{shoulder} of the histogram \cite{sahoo}. The histogram concavity analysis method is an extension of the mode method that considers points of concavity{---}valleys and shoulders{---}as
threshold possibilities.

\section{Binary Image Operations}
\hspace{0.4in}

A binary image is an image in which each pixel is one of two values{---}usually displayed as white and black. Logical and morphological processing of binary images is common. Standard boolean operations{---}NOT, AND, OR, XOR{---}are defined between pixels of corresponding locations in two images of equal dimensions, with the exception of NOT, which requires only one image. Morphological operations concern region shape. Closely related to mathematical morphology, morphological operations extract image components ``that are useful in the representation and description of region shape, such as boundaries, skeletons, and the convex hull'' \cite{russ}. Morphological operations are useful for disconnecting, thinning, thickening, and pruning objects. Morphological operations are most useful on binary and grayscale images \cite{russ}.

Open is a morphological operation so named for its tendency to separate joined objects{---}e.g. open the image. An open operation is comprised of one erosion followed by a dilation. An erosion turns edge pixels{---}black pixels with at least one white neighbor{---}white. Dilation performs the opposite operation: it turns white pixels bordering black pixels black unless doing so joins two disconnected regions of black pixels.
Each of these methods may be expanded to accept a coefficient parameter, $n$, which represents the minimum number of pixels of the opposite color that must border the pixel of interest to change its color from black to white or white to black. Thus, in an open, only those black pixels with at least $n$ black neighbors will change to white. A traditional open has a parameter of one. %This may be unclear.

\section{Color Models}
\hspace{0.4in}

Color models (also known as color spaces or color systems) define a standard way to specify a certain color. Many color models exist and each is useful for different purposes. Among the most common are RGB (used in LCD's, monitors, and popular image file formats including JPEG and TIFF), YIQ (used for color television broadcast), and CMY and CMYK (used for printing).

The RGB (Red, Green, Blue) color system defines each color within its color space using a 3{}-tuple. The 3{}-tuple specifies the percentage of red, green, and blue associated with a given color. RGB is an additive color system. When the maximum amount of red, green, and blue light are present (1,1,1), white pixels result. Black, on the other hand, is given by the absence of red, green, and blue light (0,0,0).

YIQ (Luminance, In{}-phase, Quadrature) color space separates an image's intensity information from its color information. Thus, intensity transformations may be easily realized on a color image in this color space. YIQ color space is closely related to the NTSC broadcasting standard \cite{russ} enabling color and monochrome television sets to display the same signal appropriately. The Y channel represents luminance or brightness. A color digital photograph may be converted to a black and white digital photograph by computing the Y channel of YIQ color space. This channel is formed by combining the RGB channels in proportion to the human eye's sensitivity to each color \cite{russ} as given by Equation \ref{eq:ychannel}.

\begin{equation}
\label{eq:ychannel}
Y=0.299\cdot R+0.587\cdot G+0.114\cdot B
\end{equation}

CMY (Cyan, Magenta, Yellow) and CMYK (Cyan, Magenta, Yellow, blacK) are based on the subtractive color system and are used extensively in color printers and copiers. Cyan subtracts red light from reflected white light. That is, when a surface coated with cyan is struck by white light, the red frequencies will be absorbed leaving only the green and blue frequencies{---}cyan. Likewise, magenta subtracts green light, and yellow subtracts blue light. If a surface is coated with cyan and magenta, both red and green frequencies will be absorbed and only blue light frequencies will be reflected back making the image appear blue. An image is easily moved from RGB to CMY color space by Equation \ref{eq:cmyrgb}.

\begin{equation}
\label{eq:cmyrgb}
\left[ 
\begin{array}{c}
C\\
M\\
Y
\end {array}
\right]
=
\left[
\begin{array}{c}
1\\
1\\
1
\end{array}
\right]
-
\left[
\begin{array}{c}
R\\
G\\
B
\end{array}
\right]
\end{equation}

CMYK is very similar to CMY but introduces another channel, K or black. The black channel represents a shade of gray that is common to all channels, thereby reducing the amount of colored ink needed when printing. Figure \ref{cmkycode} shows how each pixel in an image is moved from RGB color space to CMYK color space.

\begin{figure}[bth]
%% position codes are b=bottom, t=top,  h=here, p=separate page
\begin{verbatim}
 //Assume c,m,y,k,r,g,b all vary on the interval [0,1]
 c=1-r;
 m=1-g;
 y=1-b;

 k = Min(c,m,y); //Select the smallest of all three numbers

 if (k==1) {
    c=0;
    m=0;
    y=0;
 } else {
    c=(c-k);
    m=(m-k);
    y=(y-k);
 }
\end{verbatim}
\begin{center}
\parbox{6.0in}
%% this goes into the figure list (which is just after the table of contents..)
{\caption[Example code to convert a pixel from RGB color space to CMYK color space]
%% this goes under the picture
{\em Example code to convert a pixel from RGB color space to CMYK color space.}
\footnotesize

\label{cmkycode}}
\end{center}
\end{figure}

%% file: chapter3.tex
\chapter{Approach and Uniqueness}
\hspace{0.4in}

This research aims to determine what combination of image manipulation techniques reliably extracts dorsal fin outlines from digital photographs with minimal user input. While the techniques are established, this aim is unique. The algorithm involves three primary stages{---}constructing a binary image, refining the binary image, and forming and using the outline yielded by the refined binary image. Unlike classic uses of segmentation for quality assurance in a factory, the use of these techniques in the field brings many new challenges. Lighting differences, scale differences, rotational and size differences, and color differences---each complicates the segmentation process. An approach to extract dolphin dorsal fin outlines with very little \emph{a priori} knowledge is sought.

Edge detection methods are popular and may be found in such graphic manipulation packages as PhotoShop$^{\rm TM}$ and Gimp; yet, edge detection methods generally identify all edges in an image. The task then shifts to identifying which edges correspond to those of the fin outline. This research first provides a rough outline through image segmentation and morphological processing techniques. Secondly, that rough outline is refined using local edge detection methods.

For this research, two distinct approaches were developed based upon the color information in the fin images. The formation of an intensity image and identification of an optimal threshold are accomplished quite differently in each approach. The refinement of the segmented image and the extraction of the fin outline, on the other hand, do not significantly differ between the two approaches.

\section{Approach 1, Standard Intensity Image}
\hspace{0.4in}

%initial scaling, specifics of thresholding, morphological processing?}

The process begins when the user imports a photograph of an unknown
fin to compare to existing, known fins. The algorithm first seeks to
establish a rough outline of the fin to define the constraint space and
boundaries within which the Canny edge detector \cite{canny} and active
contour \cite{kass} operations may later operate. In order to find this
approximate outline efficiently, the image is first downsampled. The
operations that follow all depend on the number of pixels in the image.
Thus, reducing the image's size decreases overall processing time.
At the same time, precision suffers. This, however,
will be corrected when the rough outline is plotted and refined using the full resolution image.

%% this is the code for a figure called "method1"
%% position codes are b=bottom, t=top,  h=here, p=separate page
\begin{figure}[bt]
\centerline
{\epsfxsize=3.5in\epsfbox{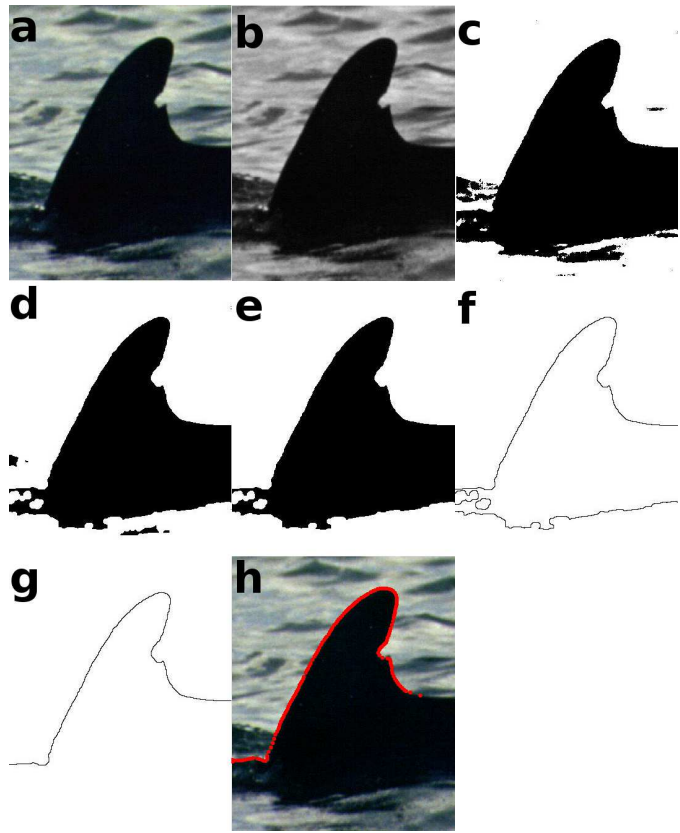}}
\begin{center}
%% this goes into the figure list (which is just after the table of contents..)
\parbox{6.0in}{{\caption[Example Progression using Approach 1]
%% this goes under the picture
{\em Intermediate steps for an automatic extraction of a fin's outline from a digital photo. (a) Color image, (b) Grayscale Image, (c) Result of unsupervised thresholding based on histogram analysis, (d) Following a series of morphological processes [open (erosion/dilation), erosion, AND with original binary image c], (e) Feature recognition to select the largest feature, (f) Outline produced by one erosion and XOR with e, (g) Feature recognition to select largest outline, (h) outline spaced/smoothed using active contours.}}
\footnotesize

\label{method1}}
\end{center}
\end{figure}

\subsection{Choosing a Threshold Value}
\hspace{0.4in}
Figure \ref{method1} illustrates the step by step approach used to extract 
an outline in a typical dorsal fin image. The color image (Figure \ref{method1}a) 
is first converted to a grayscale image (Figure \ref{method1}b)
using Equation \ref{eq:cmyrgb} so that the
intensity of each pixel may be analyzed. A histogram of the grayscale
image is constructed and analyzed. In order to keep the algorithm
general and enable it to handle a wide variation of images, only global,
point{}-dependent thresholding methods were surveyed. Point{}-dependent
algorithms make no assumption about the location of the foreground
object. Further, only global algorithms were considered as each image
is assumed to consist of exactly one dorsal fin. The p{}-title method
\cite{sahoo} was explored and discarded in order to avoid fixing a percentage of
the image that the dorsal fin must occupy, thereby limiting the domain
of images upon which the algorithm would succeed or imposing additional preprocessing tasks on the user. While optimal images
are nearly bimodal and work well with the mode method \cite{sahoo}, the
histogram concavity analysis method \cite{sahoo} was found to work well on a
larger spectrum of images and was, thus, employed. Nevertheless, in
selecting a threshold, some assumptions were necessary. The algorithm
assumes that the image consists of two regions{---}object or fin and
background. It is further assumed that the fin is the darkest of these
two regions. Using these limited assumptions, the algorithm computes
the histogram and finds the first ``valley.'' To identify the first
valley, the relative shape of the histogram is analyzed from left to
right. A valley must be proceeded by a increase and a peak or plateau.
At each position in the histogram, the count of pixels in a symmetric
neighborhood of 30 intensity values is used to determine if
the number of pixels at each intensity position is increasing,
decreasing, or level. Pixel counts within 2\% of the
count of the target position are considered to be level. After the
first valley is found, the second valley is also determined. The
distance from the end of the first valley to the tip of the second
histogram region is computed. If they are significantly close, then the
fin is likely two{}-toned, and the optimal threshold is at the second
valley.

\subsection{Fin Refinement and Outline Generation}
\hspace{0.4in}
The algorithm constructs a binary image (Figure \ref{method1}c) in the second stage
by thresholding the grayscale image at the threshold value chosen in
stage one. Morphological processes are applied to the binary image in
the third stage to produce a cleaner outline (Figure \ref{method1}d): often,
thresholding will include dark portions of waves or shadows as part of
the fin. The image is iteratively opened{---}erosion followed by
dilation{---}to separate the fin from these small dark regions. An
efficient implementation was found by adopting
Cychosz's thinning algorithm, which uses a neighborhood map approach \cite{cychosz}. After several
standard erosions, the image undergoes a series of erosions with a high
coefficient of 5. These erosions essentially remove any pixels that do
not touch at least five other object pixels. The image then undergoes a
number of dilations equal to the number of standard erosions, and the
image undergoes an AND with the binary image before any morphological
processing was done. This ensures the dilations do not add any new
object pixels and helps maintain the shape of the fin. The largest blob
of black pixels is then selected in the fourth stage (Figure \ref{method1}e).
%(Efficiency or lack thereof remark?)

The outline is formed in stage five by eroding a copy of the image once
and performing an exclusive-or (XOR) with the binary image as it existed before the erosion.
This results in a one-pixel wide outline (Figure \ref{method1}f). The
largest outline is then selected using the same code that selected the
largest blob (Figure \ref{method1}g). This is necessary as smaller
outlines often result from sun{}-glare spots on the fin and other
abnormalities.

\subsection{Outline Walk and Refinement}
\begin{figure}[bt]
\centerline
{\epsfxsize=2in\epsfbox{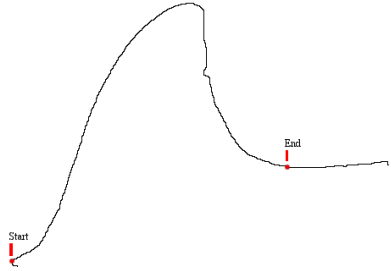}}
\begin{center}
\parbox{3.0in}
%% this goes into the figure list (which is just after the table of contents..)
{\caption[Automatically Calculated Start and End Points]
%% this goes under the picture
{\em The automatic detection of start and end points was implemented in initial software versions, but was later superseded by the refinements discussed in Section \ref{sec:refinements}.}
\footnotesize

\label{endpoints}}
\end{center}
\end{figure}

\hspace{0.4in}
In the next stage, the start and end of the fin outline
are detected and walked to form a chain of coordinate pairs. The
one{}-pixel wide outline often includes the peduncle and other portions
of the dolphin's body. A start point is
identified as the first of three points that move sequentially in a
north{}-east direction (assuming movement towards the top of the image
is defined as north). Similarly, an end point is identified as last
of three points to move sequentially in a south{}-east direction.
Figure \ref{endpoints} shows an outline with the start and end points
identified. As the outline is walked, movement is prioritized in a
east{}-north{}-south{}-west manner. That is the algorithm will first
seek to move one pixel east. If this is not possible then north
movement is tried. Only if these two directions fail is south movement
sought, etc. Some fin outlines are closed, and this priority of
movement ensures the fin is always walked left to right, start to
end.

%% this is the code for a figure called "endpoints"
%% position codes are b=bottom, t=top,  h=here, p=separate page

In the final step (also performed for manually{}-traced fins), the
algorithm plots the points on the original color image and
spaces/smooths the outline using active contours and a Canny edge
detector to reposition the outline along the true dorsal fin edge.
(Figure \ref{method1}h).

\section{Approach 2, Cyan Intensity Image}
\hspace{0.4in}

Although many dorsal fin images exhibit a good contrast between the fin
and the background when the standard intensity image is used, others,
such as those shown in Figure \ref{rgb_gray_cyan} b,e,h, do not. For
such images, an alternative representation of the images, which 
facilitates the separation of fin from background was sought.

\subsection{Choosing an Alternative Color Space}
\hspace{0.4in}

In order to
determine the optimal mapping from color space into an intensity image,
several color systems were examined comparing all possible weightings
of each channel, out to two{}-decimal places. Approach 1 was applied to
302 images and failed to produce a usable outline for 96 images. This
failure resulted despite the fact that for many of the images, the fin
was easily distinguishable from the background for a human
observer{---}using color information. In order to identify a color
space in which a suitable threshold could be easily identified for such
images, several transformations to alternative color spaces were
examined. These color systems were analyzed to determine the optimal
mapping from the original color images into intensity images, comparing
all possible weightings of each color component channel, out to
two{}-decimal places.

%% this is the code for a figure called "rgb_gray_cyan"
%% position codes are b=bottom, t=top,  h=here, p=separate page
\begin{figure}[hbt]
\centerline
{\epsfxsize=4in\epsfbox{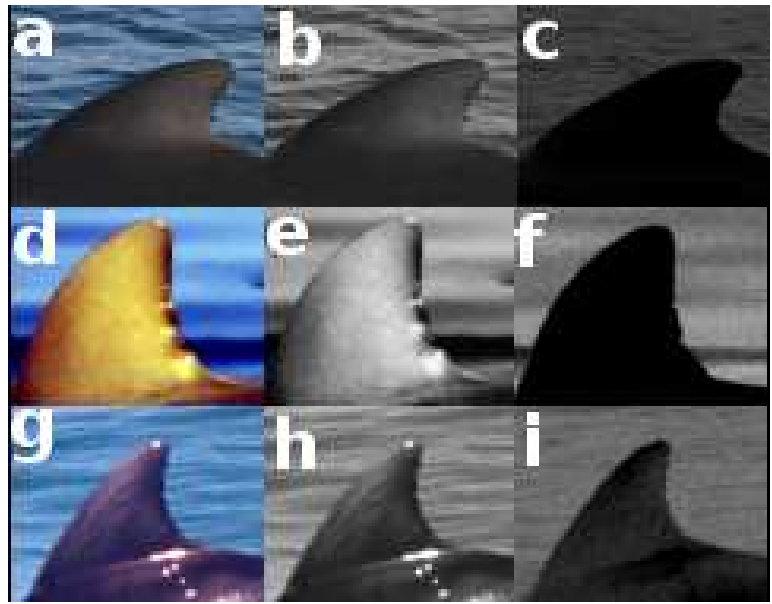}}
\begin{center}
\parbox{4.0in}
%% this goes into the figure list (which is just after the table of contents..)
{\caption[RGB, Standard Intensity Images, and Cyan Channel Images Compared]
%% this goes under the picture
{\em Three dorsal fins in RGB color space (left column), with their standard intensity images (center column), and the cyan channel of CMYK color space (right column).}
\footnotesize

\label{rgb_gray_cyan}}
\end{center}
\end{figure}

Ten images representative of the 96 were used for initial evaluation of
possible color channel weightings. For each image, an ``ideal'' image
separating the fin and background was created manually. An iterative
algorithm then compared the image in all color spaces determining which
color space and weighting classified the most pixels (fin and not fin)
correctly as determined by the manual segmentation. The best intensity
image for this representative set was formed by converting the image to
CMYK color space and using the cyan channel in place of the original
intensity image for the purposes of thresholding (Figure \ref{rgb_gray_cyan} c,f,i). For
this set of images, the cyan channel yields good contrast between the
cyan{}-rich water and cyan{}-poor fin in many cases.

\subsection{Iterative Unsupervised Thresholding}
\hspace{0.4in}

The intensity images produced using the cyan channels of the original
images are then subjected to an iterative thresholding algorithm.
Traditional histogram analysis methods \cite{sahoo}, whether point{}- or
region{}-dependent and global or local, do not consider the arrangement
of the pixels in evaluation of the threshold. In order to achieve a
useful portioning of the image into object and background, a technique
for analyzing the resulting arrangement of object and background pixels
was sought.

A ``pixelarity'' metric was developed to evaluate the quality of the
segmentation produced using a particular threshold value. The
evaluation is based upon the degree to which a threshold produces large
solid areas of fin and background. This method requires limited
\emph{a priori} knowledge, unlike other traditional histogram
analysis methods which require information about the size (p{}-tile
method \cite{sahoo}), location, etc., of the feature of interest.

\subsection{Evaluation of Threshold Quality}
\hspace{0.4in}

To evaluate the quality of a threshold, a three by three structuring
element is passed over the thresholded image and a score is assigned to
each arrangement of pixels the element encounters. Then, the average
score for the image is computed.
A given pixel arrangement is scored using a neighborhood map approach
\cite{cychosz}. The arrangement of the nine pixels beneath the structuring element
is mapped one{}-to{}-one with a number in the range zero to 512. A look
up table holds the score for each possible arrangement of pixels. 

\begin{figure}[h]
\centerline
{\epsfxsize=1in\epsfbox{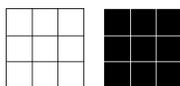}}
\begin{center}
\parbox{6.0in}
%% this goes into the figure list (which is just after the table of contents..)
{\caption[Best Case Pixel Arrangements]
%% this goes under the picture
{\em The best case arrangements are large solid areas.}
\footnotesize

\label{best_case}}
\end{center}
\end{figure}

\begin{figure}[h]
\centerline
{\epsfxsize=1in\epsfbox{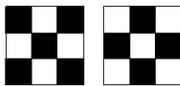}}
\begin{center}
\parbox{6.0in}
%% this goes into the figure list (which is just after the table of contents..)
{\caption[Worst Case Pixel Arrangements]
%% this goes under the picture
{\em The least desirable arrangements are highly fragmented areas.}
\footnotesize

\label{worst_case}}
\end{center}
\end{figure}

Ranging from one to 21, the score each arrangement of pixels receives is
the sum of the number of long edges and the number of four{}-connected
objects in the region beneath the structuring element, as described in Section \ref{sec:pixmetric}. 
For example, the best score, one, is achieved
with either of the cases (one region with no edges) shown in Figure \ref{best_case}.
The least{}-desirable score, 21, is achieved with either of
the ``checkerboard'' patterns (nine regions and 12
edges) shown in Figure \ref{worst_case}.
%NEED to insert section cross reference above

\begin{figure}[bt]
\centerline
{\epsfxsize=5in\epsfbox{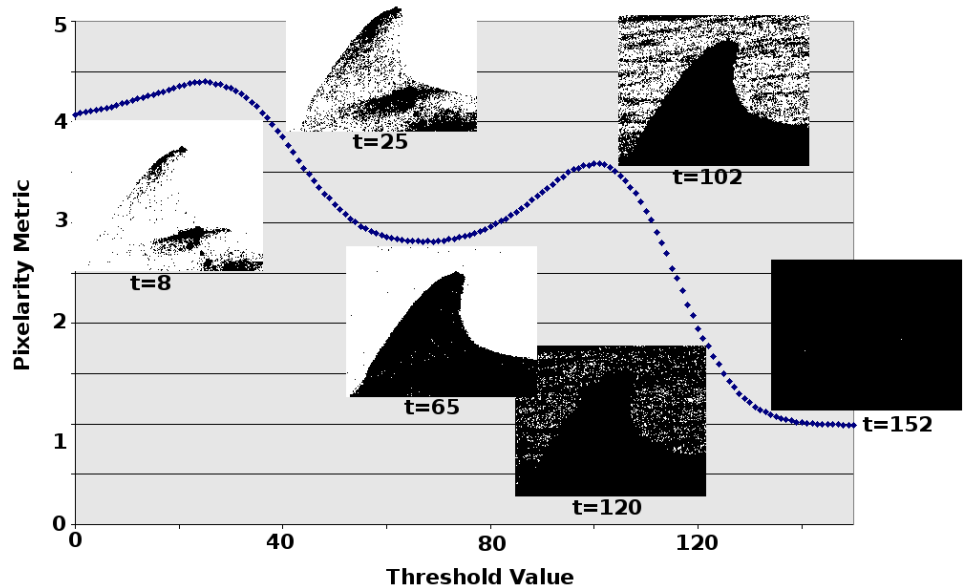}}
\begin{center}
\parbox{6.0in}
%% this goes into the figure list (which is just after the table of contents..)
{\caption[Pixelarity Graph]
%% this goes under the picture
{\em The ``pixelarity'' score (y-axis) as the threshold varies from 0 to 160 (x-axis).
Several thresholds are illustrated along the graph.
The threshold 65, located at the center local minimum, is the ideal threshold as selected by the algorithm.}
\footnotesize

\label{graph}}
\end{center}
\end{figure}

Increasing thresholds are taken in increments of five and
the ``pixelarity'' of each is computed. Figure \ref{graph} shows the graph of the
``pixelarity'' of an image as the threshold increases. The pixelarity
graphs of images with two distinct regions (object and background)
share a similar shape. At first the ``pixelarity'' increases as the
darker region is selected. Pixelarity then decreases, and the graph
reaches a local minimum, representing the ideal threshold. Past this,
pixelarity again increases as pixels from the lighter region become
classified as part of the darker region. With significantly high
thresholds, both object and background are selected and the pixelarity
measure approaches one (a solid{}-colored image). The grayscale images
of dolphin dorsal fins formed from the cyan channel in CMYK
color{}-space are subjected to this pixelarity measure and thresholded
at the first local minimum.

%% this is the code for a figure called "graph"
%% position codes are b=bottom, t=top,  h=here, p=separate page

\subsection{Morphological Processing and Walk}
\hspace{0.4in}

Once thresholded, an outline is produced from the image
using steps similar to those employed in Approach 1. Open and close
morphological operations clean up the fin edge, and a one{}-pixel
erosion on a copy of the image is used in an exclusive-or (XOR) with
the thresholded image to produce a one{}-pixel wide outline of all
objects. The largest outline is selected and walked to form a chain
code to be used as a digital representation of the fin outline for
comparison operations.

\subsection{Developing the ``Pixelarity Metric''}
\label{sec:pixmetric}
\hspace{0.4in}

The score for a given 3x3 arrangement of pixels is the sum of the
number of distinct four{}-connected objects and the number of long
edges. The scores range from one (1) to twenty{}-one (21). In
developing this metric, several alternatives were evaluated.

\begin{figure}[h]
\centerline
{\epsfxsize=1in\epsfbox{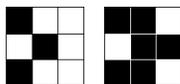}}
\begin{center}
\parbox{6.0in}
%% this goes into the figure list (which is just after the table of contents..)
{\caption[Four- vs. Eight-Connected Region Comparison]
%% this goes under the picture
{\em A four-connected definition of regions produces more accurate pixelarity scores
than does an eight-connected definition.}
\footnotesize

\label{four_eight}}
\end{center}
\end{figure}

In determining the number of objects, eight{}-connected and
four{}-connected approaches were considered. Each pixel in a
four{}-connected object must touch another pixel in the object directly
above, below, left, or right of itself. Eight{}-connected objects offer
a more lenient definition of connectedness, and pixels in such an
object need only touch one other pixel of the object in any direction,
including diagonals. Thus, all four{}-connected objects are
eight{}-connected, but not all eight{}-connected objects are
four{}-connected. Using eight{}-connected objects in place of
four{}-connected objects yields scores that are unrealistically
favorable. In a well{}-segmented fin image, we seek large solid
elements. The fin should be well{}-connected using four{}-connected
objects. Using a the four{}-connected definition, various isolated
diagonal segments are penalized more strongly for not having any other
pixels connect them.
The pixel arrangement on the right of Figure \ref{four_eight} simply selects three more pixels as object than does
the arrangement on the left. Yet, this addition of pixels raises the pixelarity score
counter to intuition when an eight-connected definition is used. 
Using an eight-connected definition in Figure \ref{four_eight}, the arrangement on the left
receives a score of 10 (two regions and eight edges) and the arrangement on the right
receives a score of 11 (four regions and seven edges).
Using a four-connected definition, the leftmost arrangement scores 13 (five regions and eight edges), while the rightmost arrangement receives a superior score of 11 (four regions and seven edges).

\begin{figure}[h]
\centerline
{\epsfxsize=1in\epsfbox{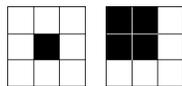}}
\begin{center}
\parbox{6.0in}
%% this goes into the figure list (which is just after the table of contents..)
{\caption[Long vs. Short Edge Comparison]
%% this goes under the picture
{\em Employing long edges breaks ties between pixel arrangements such as these.}
\footnotesize

\label{long_short}}
\end{center}
\end{figure}

In counting the number of edges in a 3x3 pixel arrangement, multiple
schemes are possible. One option is to count all edges{---}that is
wherever a white and black pixel meet could be considered one edge. This is referred to
as the number of ``short edges.'' Long edges, on the other hand,
consider the surrounding pixels. Where one short edge is the natural
extension of another (connecting at one end point and continuing in the
same direction) these two short edges can be thought of as producing
one ``long edge.'' Using long edges gives scores that more closely
parallel the intuitive evaluation of a pixel arrangement than does
using short edges. The use of long edges gives preference to straight
clean divisions between foreground and background. Long edges also
result in fewer ties between scores for pixel arrangements. For
example, the arrangement on the left in Figure \ref{long_short}, scores six (two regions and four
edges) using either long or short edges. The arrangement on the right in Figure \ref{long_short},
which seems intuitively superior to the one on the left, receives the
same score of six (two regions and four edges) employing short edges.
This tie is broken using long edges, which score the arrangement on the right as four (two regions and two long edges).

%FIGURE long_short.ps Employing long edges breaks ties between pixel arrangements such as these.

The pixelarity scores for each 3x3 neighborhood are totaled and the mean
score is calculated. This mean score is highly dependent upon the
image's contents and cannot be compared from one image
to another. There is no mean score that indicates a proper threshold;
rather, the relation between the scores of differing thresholds offers
clues as to the ideal threshold.

Through analysis of a sample set of 78 images that failed to extract a
usable outline under Approach 1, three categories of images emerged.
These categories were established based on the graphs of the mean
pixelarity scores at iterative thresholds. One category consisted of
images with a concave{}-up portion in their graphs, another category consisted of
images with a plateau in their graphs, and the final category consisted of
images with no critical points in their graphs.

The first category has a shape akin to Figure \ref{graph}. The ideal
threshold is the first local minimum. At smaller thresholds, nearly all
the image is selected as background, and the pixelarity score is low.
As the threshold first increases the pixelarity score rises as
foreground and background elements separate. Soon the graph begins to
decrease as foreground and background become better segmented. At the
local minimum, the ideal threshold is reached. Past this point, higher
thresholds begin to select background elements, and the pixelarity
score again rises until the threshold begins to select the majority of
the image. As this happens, the pixelarity score falls and eventually
approaches one. A pixelarity score of one is achieved when the entirety
of the image is selected as foreground or background. The ideal
threshold for (a bimodal image with) a graph of this shape is the first
local minimum. It is precisely at this point that foreground and
background are separated most cleanly. In the second category of
images, a local minimum does not exist. Rather, a critical point is
found on the plateau of the graph, and this point represents the ideal
threshold.

In the above two categories, the ideal threshold is found among the
graph's critical points{---}where the
graph's first derivative is zero. For the graphs of a
small subset of the images, no critical points exist. In these cases,
the pixelarity score usually decreases uniformly as the threshold level
increases. In this case, the ideal threshold may be at zero (0), or the
graph may not indicate the ideal threshold. The algorithm proceeds with
an initial threshold of zero. If later operations{---}threshold
refinement, outline generation and walk{---}fail or if the final outline
produced is unreasonably short, then it is likely that zero was not the
correct threshold. In this case, the user will be asked to supply a
manual trace.

\section{Refinements to Both Approaches}
\label{sec:refinements}
\hspace{0.4in}

After development of the above two approaches, results from independent
testers using larger image sets distinct from those used in development
led to a number of observations and changes.

Four principal changes were made. These changes are 1) the user
must indicate the start and end of the fin prior to fin extraction, 2)
the fin is automatically cropped before beginning the threshold
algorithm, 3) the walk of the one{}-pixel outline was rewritten, and 
4) the largest feature code was rewritten in a more efficient manner.

Analysis showed that, occasionally, the algorithms were unable to find a
suitable starting point to walk the one{}-pixel wide outline.
At times, flipping the fin horizontally would yield a suitable starting point. 
Even so, users seemed to disagree about the exact start and end points
of the fin. As a direct response to user requests, a ``cleaver'' tool
was initially added to chop off all the points before or after an
indicated point. Thus, users were observed engaging in the following
interaction process: a) autotrace the fin b) select the cleaver tool c)
indicate the start of the fin (removing all points before the indicated
point) and d) indicate the end of the fin (removing all points
following the indicated point). By refining this process to indicate
the start and end of the fin before attempting autotrace, the algorithm
can provide a superior autotrace and do so in a more efficient manner.
Best of all, this change truly adds no complexity or user time as users
were already indicating the start and end of the fin with the
``cleaver'' tool post{}-trace. The order of the steps is simply
rearranged. Users must now indicate the start and end of the fin. As
soon as the end of the fin is indicated, the program will attempt an
autotrace. 

With the additional information of the start and end points of the fin
available, the fin may be automatically bounded. By focusing the global
threshold algorithms on the specific region of the image containing the
fin, excessive noise in the histogram is reduced and the chances of
achieving a successful outline increase. This automatic bounding
process was attempted in two ways. In the first attempt, after the
existing scaling operations, fins were cropped to a bounding box
formed from the user start to end points plus 100px padding and a
height of one and a half times the width. Such height (based upon the
golden ratio) worked well in a sample of photographs. Under this
approach, scaling was desirable so that set values could be used to
pad the fin. However, by scaling extremely large images before cropping, the
fin became too small to achieve the necessary accuracy. Thus, a second
approach was employed. Users were again observed zooming images so that
small fins occupied the majority of the view area. Such action is a
necessary precursor to verifying an autotrace or providing a manual
trace. The algorithm now bounds to the viewable area. This introduces
one major human{}-computer interaction issue: the entirety of the fin
must be visible in the view area if the algorithm is to succeed. This
information must be aptly and nonintrusively communicated to users.

%How to address this communication issue?

The code to walk the outline was rewritten. All outlines are
four{}-connected, and they are now walked as such. This leads to fewer
decision points where multiple directions are possible. In addition,
user supplied start and end points are now used to better walk the
outline. In the sequence of points formed by walking the fin, order is
important. Previously, the exact start and end points of the outline
were difficult to determine. Even with a user{}-supplied start point,
the closest point on the outline to that point must be determined. At
first an expanding search pattern centered at the user{}-start point
was used, and the outline was walked from the first point encountered.

\begin{figure}[h]
\centerline
{\epsfxsize=3.5in\epsfbox{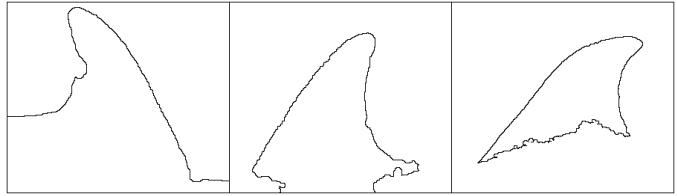}}
\begin{center}
\parbox{4.0in}
%% this goes into the figure list (which is just after the table of contents..)
{\caption[Various Outline Shapes Formed]
%% this goes under the picture
{\em Both closed and open outline forms result from the automated extraction process.}
\footnotesize

\label{outline_categories}}
\end{center}
\end{figure}

To ease this situation and ensure the optimal point is used, a new
approach was developed. Under this approach, a pseudo{}-secant line is
first calculated between the start and end points. This line is the
minimum y component of the points and runs parallel to the x{}-axis for
the length of the distance between the two points. The perpendicular
bisector to this line (parallel to the y{}-axis) is then presumed to
intersect the fin outline one or more times. Theoretically, two 
outline categories exist (Figure \ref{outline_categories}). Where there is
sufficient contrast between the fin and the dolphin's
body, a closed{}-form outline often results. In other cases, an
open{}-form outline results and touches various sides of the image. The
closed{}-form outline presented the greatest challenge for the
expanding search approach. Under that approach, the bottom of the fin
was sometimes first encountered and then the fin was walked from end to
start. Now, the north most point to touch the vertical bisector is used
as a seed point. The algorithm first walks to the east of this point as
far as possible. As the eastward walk progresses, the algorithm makes
certain never to cross the perpendicular bisector. This  check
allows closed{}-form outlines to be treated identically to open{}-form
outlines. This check may be thought the equivalent of inserting a small
one{}-pixel gap on the line formed between the
dolphin's body and fin in such closed{}-form outlines.
After the eastward walk concludes, the algorithm returns to the seed
point and walks westward prepending each point encountered. The user
start and end points are then used to select the coordinate pairs in
the outline that are closest to these points as calculated by the
standard Euclidean distance formula. The sequence of points is trimmed
and reversed as necessary.

%FIGURE outline_categories (closed, open)

The ``largest feature'' code was the most time{}-intensive and problematic phase
in both Approach 1 and Approach 2. During this stage, the image is traversed row by row to 
identify connected ``blobs'' of zero-valued pixels, the largest of which is assumed to be the fin.
The preprocessing step of scaling
the image reduced the number of pixels to be evaluated, but
the upper limit on the number of possible blobs
(connected groups of black pixels) encountered is still quite large, $n/2$ if the image has $n$ pixels. Images with poorer
thresholds often contain much noise, and the number of blobs in such images
is very high. In one case the number of blobs was so high so as to
cause the algorithm to terminate abnormally. These excess blobs are
generally located in the water and other non{}-central parts of the
image. With the start and end point of the fin known, the search area
for blobs can be greatly reduced. A point is identified with an $x$ coordinate
that corresponds to the midpoint of the pseudo{}-secant line connecting the
user-selected start and end points. The $y$ value of this point is equal to
the $y$ value of the midpoint minus one-half the length of the pseudo-secant line.
Ostensibly, the feature which contains this point should be the fin.
Theoretically, however a slight abnormality
at this exact point could cause an error. Such an abnormality could
result from severe discoloration, sun glare, etc. While the search for
multiple blobs cannot be completely abandoned, the search range can be
significantly reduced from the whole image to a small region around
this point. This reduction results in large computing time{}-savings and speeds
both approaches.

Finally, the knowledge of user{}-defined start and end points assist in
the evaluation of an automatically produced outline. If the outline is
unreasonably short or does not approach the start and end points adequately
 the outline is discarded. This evaluation is
important in the current multiple tier implementation. Effective
evaluation allows the use of multiple approaches without user knowledge
or further interaction.

%% file: chapter4.tex
\chapter{Results and Contributions}
\hspace{0.4in}

Work performed as a part of this thesis research demonstrated methods to successfully extract an outline of a dolphin dorsal fin in a large percentage of images acquired in the field. The automated fin extraction process reduces user input time and eases
visual fatigue, improving the end user experience. 

\section{Approach 1, Standard Intensity Image \\(Before Modifications)}
\hspace{0.4in}

The Approach 1 was developed using a set of 36 images collected
from field research. The test set of images for this approach included 302 images
distinct from the 36 in the development set. These 302 images were
collected as part of an active population study of bottlenose dolphins
(\emph{Tursiops truncatus}) in Boca Ciega Bay, FL. In the set, the program
automatically produced a usable outline for 106 images (35.1\%).
Another 100 images (33.11\%) required only slight modification to the
automatically extracted fin outline. The autotrace algorithm was unable
to produce an outline for 78 images (25.83\%). Finally, the algorithm
produced an unusable outline (usually selecting the wrong feature) for
18 images (5.96\%).

Two{}-thirds of the images traced successfully{---}half with no
modification whatsoever by the user. This significantly reduces the
time and visual fatigue in fin outline extraction and greatly
facilitates use of automated recognition packages. Given the large
quantity of photographs to be compared, avoiding a manual tracing
process for two{}-thirds of the images greatly reduces end{}-user time and fatigue.
Reactions expressed by marine scientists at the 16th Biennial
Conference on the Biology of Marine Mammals suggest this research has
the potential to improve user experience with automated recognition
packages. In the event of failing to identify the outline, a user
traces the outline as usual with no loss in time; while, in the case of
a successful extraction of an outline, the user proceeds directly to
matching the fin, bypassing the time{}-consuming and
visually{}-fatiguing manual tracing process.

%FIGURE Graph of results for Approach 1

\section{Approach 2, Cyan Intensity Image}
\hspace{0.4in}

The second approach was developed based on a sample of images
which failed Approach 1. A testing set of 94 images was formed from
images that failed to produce usable outlines with previous approaches.
This set was distinct from the 96 images originally used in developing
this approach. Of the 94 images tested, 48 (51\%) of these
automatically produced a successful trace using Approach 2.

The metric scores, as applied to thresholding, are relative to a
particular image and cannot be used to compare thresholds across
multiple images. Overall, however, the metric provides information
about pixel arrangement and mimics the way a human chooses a threshold.
The method provides good results even where traditional histogram
analysis methods prove inconclusive, and this method further requires
limited \emph{a priori} knowledge. The algorithm
runs on the order of $O(n)$ overall where $n$ is the number of pixels in
the image. The metric may also be useful in evaluation of image
compression algorithms by comparing ``pixelarity'' measures of an image  %%KRD asks how??
before and after compression.

\section{Comparison}
\hspace{0.4in}

Both approaches use similar steps for outline production, refinement,
and walking. The computation of the standard intensity image and the computation
of the cyan intensity image both incur similar processing expenses. The
largest difference in runtime stems from the thresholding stages of the
processes. Both approaches are of linear complexity, $O(n)$, relative to
the number of pixels in the image; yet, Approach 2 still incurs a
greater runtime expense as multiple threshold{}-images must be formed
and analyzed. As Approach 1 determines the threshold from the
histogram, only one threshold{}-image is produced.

Given the increased runtime of Approach 2, the software
implements the two algorithms in a tiered approach. Only those images
which fail the first approach are subjected to the second process.
Approach 1 yielded a usable outline for 68\% of images. If the
testing set for Approach 2 is representative of the 32\% failing
previous approaches, then a tiered solution combining both approaches
ought to produce usable outlines for approximately 84\% ($68\%+51\% \cdot 32\%$) of a
typical set of dorsal fin images.

In testing with 224 images which failed Method 1 before the
modifications of Section \ref{sec:refinements}, 130 of the images (58\%) now produced a
usable outline with little or no user modification. Using a tiered
approach with Method 1 and Method 2, 178 images (79\%) produced usable
outlines. If this set is representative of the 32\% failing Method 1
before modifications, then the modifications and tired solution
together should raise the overall percentage of usable outlines to
approximately 93\% ($68\%+79\% \cdot 32\%$) of a typical set of dorsal fin images. It should be
noted that the test set of 94 images used to test Method 2 was a subset
of these 224 images. Testing of a general set with the tiered approach
is ongoing, and no large scale results are yet available.

Prior to implementations of autotrace algorithms, users needed to hand
trace all images. With a tiered solution, the number of images
requiring a manual trace is significantly reduced. This reduction
should improve usability, ease visual fatigue, and decrease the time{}-consuming nature of 
data entry for automated photo-identification.

%[REF]Handtrace accuracy vs. auto trace=good enough

%Compare Approach 1 to Approach 2 (complexity, runtime, result success rate)

%Tiered approach with both (avg weighted run time, overall success)

%% file: chapter5.tex
\chapter{Usability Concerns / HCI}
\hspace{0.4in}

The development of fin-outline extraction algorithms is closely related to the field of Human Computer Interaction (HCI). The goal of such algorithms is to improve the user experience and the usability of fin-comparison software packages. Yet, these goals cannot be met without proper use of the algorithms within the software package. Thus, HCI research must guide how such algorithms are integrated into software packages and how they appear to users. This section will focus on the steps taken to modify DARWIN while adding AutoTrace functionality.

\section{Intuitive Design}
\hspace{0.4in}
The addition of AutoTrace functionality sought to be as intuitive as possible. Users should not be expected to know that a tiered approach of multiple algorithms exists. At the user level of abstraction, there is one feature---AutoTrace. It should not concern the user how this functionality is implemented at lower levels of abstraction. For a tiered approach to be successful while at the same time transparent to the user, it is important that the autotrace algorithms recognize when each fails to produce a valid outline. If Approach 1 fails, the second approach may be tried seamlessly with no user action, and if this second approach fails too, the user is guided to provide a manual trace.

Users may not understand the value of cropping an image in regards to the algorithm, nor, should this be expected. The technical benefits of cropping an image are clear: limited search space, reduced histogram noise, decreased overall processing time, etc. The target user, a biologist, should not be expected to weigh such technical considerations. Thus, it is desirable to automatically bound the image without overt user action. To accurately determine the start and end points of the fin outline, users intuitively zoom in on the region of the photograph that contains the fin. The implementation capitalizes on this approach to use the viewable area of the image as an automatic bound. The danger in doing so is that the entire fin must be available. The user should be alerted to this in several ways: on screen in the status bar, in any error dialogs should AutoTrace fail, and in user documentation.

Intelligent tool selection is also an important user interface feature so that the user interface makes the transition from task to task ``a seamless, transparent, and even pleasurable experience'' \cite{watzman}. After a successful AutoTrace, the eraser tool is selected automatically to allow for slight refinements. If AutoTrace fails, the pencil tool is automatically selected to prepare the user for a manual trace.

%\section{Cleaver Tool vs. Start and End Point Selection}
%\hspace{0.4in}

%cleaver vs. start/end
%Start/end point selection = minimal (really already provided because of field disagreement)

\section{Serial Process}
\hspace{0.4in}

Tracing a fin is a multistep serial process: image adjustments, fin trace, feature identification. Each step in the process depends upon completion of previous steps. The trace window in DARWIN uses three side-by-side buttons to refer to these three steps. Previously different buttons would be hidden or shown depending upon the current and previous stage of the process. These unexplained changes in the display of buttons could potentially cause confusion for the user.  Now, these three buttons are always shown. The current stage is indicated, and the user can see the whole process as well as his or her progress through the process. The buttons are also numbered 1, 2, 3 to designate that each stage occurs in order. %(See Figure \ref{tracewin}).

Such navigational assistance is important. The serial process may be equated to progress through a novel. When reading a book, the reader sees his or her progress through the book. The user has the same desire to know where he or she is in relation to the whole process when using an application \cite{watzman}. For this reason, all three buttons of the process are always visible. Even though the user cannot skip from the first step to the third, it is beneficial to the user to know that three steps exist and that he or she is progressing towards this final step.

At times, serial processes are modeled with ``Back'' and ``Next'' buttons. This is often used, for example, in installation programs. For the task faced in DARWIN, numbered buttons are superior to this approach. The numbered buttons provide an overall sense of the task and the user's status within the process.

%\section{Visual Design}
%\hspace{0.4in}

%cursor, icon design

\section{Documentation}
\hspace{0.4in}

Documentation should be given equal importance to other implementation concerns. Documentation includes all materials that support the user and help him or her ``achieve goals and accomplish tasks'' within a software system \cite{mehlenbacher}. This broad definition embraces not only out-of-application support items such as readme files and user guides, but also in-program support such as error message dialogs and status indicators which are integrated into the interface.

%\subsection{User Guide}
%\hspace{0.4in}
Users may consult a user guide briefly upon being introduced to a system. After this, most users do not return to it until they encounter a problem. At this point, users skip, scan, and skim documentation looking for a the solution to their specific need. While documentation is recognized as providing two types of information: procedural information (task focused, how to use an application, etc) and declarative information (how the software works, conceptual details), the majority of information should be task-focused \cite{mehlenbacher}.

The documentation for DARWIN was mostly software focused. It analyzed the program window by window and offered descriptions of each tool available in the window. While a description of each button may be valuable to some, a user looking to accomplish a certain task does not wish to read the description of each button until he or she arrives at the button which accomplishes this task. New task focused sections were added to the documentation to assist the user.

%\subsection{In Program Documentation}
The program faced an important documentation process in communicating the steps for an AutoTrace to the user as clearly, but nonintrusively, as possible. An icon and cursor were developed for AutoTrace functionality. The icon and cursor build upon the ``magic wand'' image found in many photo editing environments. The goal is to communicate to the user that the computer will complete the trace. A message in the status bar asks the user to ``Click the start and then end point of the fin.'' The program attempts to catch common errors, such as clicking the end point before the start point. The dolphin is known to be swimming to the user's left; so, if the end point is to the left of the start point, a dialog immediately alerts the user to click the start point before the end point and remember that the dolphin must be swimming to the user's left. The start and end points are automatically cleared for the user to select these points. %We probably could just reverse these two and continue without further botherment to the user. No?
Other error messages within DARWIN were updated and expanded to reflect program updates and to better inform users of the probable cause of the error. 

%Mehlenbacher

%Correct error dialogs
%Documentation: More task focused (previously focused on window). Updated, formatted.

%Shift-click for cyan autotrace only?? More testing tool?

\section{Future HCI Tasks}
\hspace{0.4in}

User observation should fuel future changes and user-interface optimizations. Among these changes will likely be further restructuring of the help files and development of more in-program assistance.

Recent observations have identified the need to have two last opened file indicators. One for traced fins and one for new images. Currently, the program remembers the last viewed file and returns the user to this file when ``Open'' is selected. However, the task of switching between opening images and opening traced fins could be made more convenient. The program returns to the last opened file (whether traced fin or image) no matter whether the user seeks to open an image or a traced fin. A superior user experience can be created by remembering the last opened image and the last opened traced fin independently. The two buttons---Open Fin and Open Image---may too closely resemble one another or be placed too close together on the tool bar as well.

Tablet computing and network distributed catalogs will likely change the way users interact with DARWIN and will require extensive HCI consideration.

%% file: chapter6.tex
\chapter{Future}
\hspace{0.4in}

\begin{figure}[h]
\centerline
{\epsfxsize=2in\epsfbox{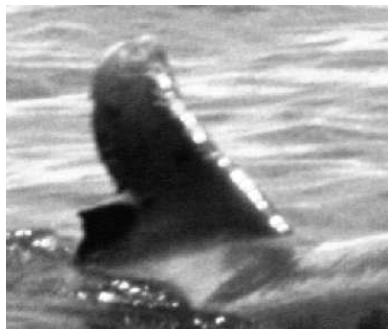}}
\begin{center}
\parbox{3.0in}
%% this goes into the figure list (which is just after the table of contents..)
{\caption[Fin Image of Poor Quality]
%% this goes under the picture
{\em Images such as this are hard to trace even manually.}
\footnotesize

\label{poor_quality}}
\end{center}
\end{figure}

The AutoTrace algorithms fails on two principle categories of images---those images images of very poor quality and those images in which another dolphin is behind the fin of interest. Those images within the first category, such as Figure \ref{poor_quality}, are often even difficult to manually trace. The prognosis for automatically tracing such images is poor.

For the second category of images---those in which another dolphin is behind the fin of interest---AutoTrace should still be possible. Using current methods, this category presents a challenge as both foreground and background are dolphin, and the two do not usually differ significantly in color. To successfully trace such outlines, this special case must be recognized and the algorithm must compensate. Often a partial outline of such fins is possible, but the outline does not approach the user end point and/or start point. The algorithms recognize this as a failure, and the outline is discarded. If this case is recognized, edge detection between the end of the partial trace and the start or end points may be able to provide a suitable trace.

\begin{figure}[h]
\centerline
{\epsfxsize=2in\epsfbox{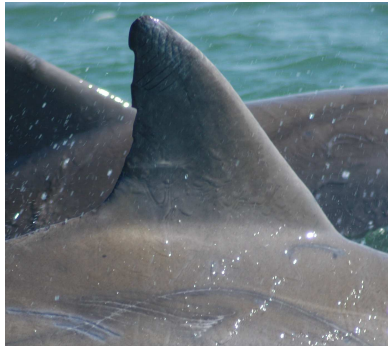}}
\begin{center}
\parbox{3.0in}
%% this goes into the figure list (which is just after the table of contents..)
{\caption[Two individuals overlapping in one image]
%% this goes under the picture
{\em The background behind the fin of interest closely matches the color of the fin and presents a problem to traditional segmentation approaches.}
\footnotesize

\label{two_individuals}}
\end{center}
\end{figure}

Bright spots are occasionally found in images where the sun reflects off of the dolphin and any water on its body. This sun glare can interfere with the outline extraction process. Preprocessing should be able to identify these glare spots and fill them based on the surrounding colors. Outline extraction may be further refined by the use of texture, mathematical modeling (spline), and edge detection. Future research may also further optimization of current approaches.

Finally, DARWIN and outline extraction, may apply to other animal species. Even where the outline itself is not the principle concern, such as whale flukes, this research may prove helpful. Identifying whale flukes involves surface characteristics primarily and not edge characteristics. Even so, this research may prove helpful in isolating and selecting the flukes from a digital photograph. In turn, the flukes may be aligned to one-another as part of a preprocessing step. These steps may greatly aid identifying such individuals.

%Multiple individuals in one photograph. (backround same color as object)
%Application to other species
%Optimization
%Use of texture, mathematical modeling (spline)

%% file: darwin-thesis.bbl
\begin{thebibliography}{100}
%required parameter is the max number of entries so the formatter knows how to align --SAH

\bibitem{canny} J. Canny. ``A Computational Approach to Edge Detection.'' In \emph{IEEE Trans. on Pattern Analysis and Machine
              Intelligence}, 8:679-689, 1986.

\bibitem{cychosz} J. M. Cychosz. ``Efficient Binary Image Thinning Using
Neighborhood Maps.'' In {\em Graphics Gems IV}, pages 465-
473. Academic Press, Inc., Cambridge, MA, 1994.

\bibitem{darwin} DARWIN. Eckerd College. darwin.eckerd.edu.

\bibitem{finscan} FinScan. Texas A\&M University.

\bibitem{gonzalez} R.C. Gonzalez and R.E. Woods, \emph{Digital Image Processing}, 2nd ed. Prentice Hall, Upper Saddle River, NJ, 2002.

\bibitem{kass} M. Kass and A. Witkin and D. Terzopoulos. ``Snakes: Active Contour Models.'' In \emph{International Journal of Computer Vision}, pages 321-331, 1988.

\bibitem{mehlenbacher}B. Mehlenbacher, ``Documentation: Not Yet Implemented, But Coming Soon!'' In \emph{The Human--Computer Interaction Handbook: Fundamentals, Evolving Technologies and Emerging Applications}, pages 527-543. Mahwah: Lawrence Erlbaum Associates, 2003. 

\bibitem{russ} J.C. Russ, \emph{The Image Processing Handbook}. CRC Press Inc., Boca Raton, FL, 1992.

\bibitem{sahoo} P. K. a. Sahoo. ``A Survey of Thresholding Techniques.''
{\em Computer Vision, Graphics, and Image Processing}, 41:233–260, 1988.


\bibitem{watzman} S. Watzman, ``Visual Design Principles for Usable Interfaces,'' In \emph{The Human--Computer Interaction Handbook: Fundamentals, Evolving Technologies and Emerging Applications}, pages 263-285. Mahwah: Lawrence Erlbaum Associates, 2003.


%original \bibitem{sahoo} P.K. Sahoo, S. Soltani, A.K.C. Wong, and Y.C. Chen, ``A Survey of Thresholding Techniques,'' {\em Computer Vision, Graphics, and Image Processing}, Vol. 41, pp. 233 - 260, 1988.

%darwin
%finscan
%hale_scott
%sahoo (already exists!)
%russ
%cychosz

%[1] DARWIN. Eckerd College. darwin.eckerd.edu.
%[2] FinScan. Texas A&M University. 
%[3] Hale, Scott. Unsupervised Thresholding and Morphological Processing for Automatic Fin-outline Extraction in DARWIN (Digital Analysis and Recognition of Whale Images on a Network). ACM Digital Library.
%\bibitem{sahoo} P.K. Sahoo, S. Soltani, A.K.C. Wong, and Y.C. Chen, ``A Survey of Thresholding Techniques," {\em Computer Vision, Graphics, and Image Processing}, Vol. 41, pp. 233 - 260, 1988.
%[5] J.C. Russ, The Image Processing Handbook, 2nd ed. 1995.
%[6] Cychosz, J. M. ``Efficient Binary Image Thinning Using Neighborhood Maps,'' Graphic Gems, IV. 465-473.
%Gonzalez
%mehlenbacher
%watzman


\end{thebibliography}
